%% file: Video_based_Heart_Rate_Estimation_with_Angle_guided_ROI_Optimization_and_Graph_Signal_Denoising/ICASSP2026_Paper_Templates/Template.tex
\documentclass{article}
\usepackage{spconf,amsmath,graphicx,hyperref}
\usepackage{enumitem}
\usepackage{multirow}    
\usepackage{booktabs}    
\usepackage{array}       
\usepackage{graphicx}    
\usepackage{amsmath}  
\usepackage{amsmath,amsfonts}
\usepackage{algorithm}
\usepackage{array}
\usepackage[caption=false,font=normalsize,labelfont=sf,textfont=sf]{subfig}
\usepackage{textcomp}
\usepackage{stfloats}
\usepackage{url}
\usepackage{verbatim}
\usepackage{graphicx}
\usepackage{xcolor}
\usepackage{cite}
\usepackage{multirow}
\usepackage{array}
\usepackage{booktabs}
\usepackage{algpseudocode}
\usepackage{hyperref}

\title{Video-based Heart Rate Estimation with Angle-guided ROI Optimization and Graph Signal Denoising}
\name{Gan Pei$^{1}$, 
Junhao Ning$^{1}$,
Boqiu Shen$^{1}$,
Yan Zhu$^{2*}$, 
Menghan Hu$^{1,3*}$\thanks{This work was supported by the National Natural Science Foundation of
China (Grant 62371189), the Natural Science Foundation Project of Chongqing (Grant cstc2021jcyj-msxmX0816) and the STCSM (Grant 22DZ2229005). $^{*}$Corresponding authors: Yan Zhu and Menghan Hu.}}
\address{
    $^{1}$Shanghai Key Laboratory of Multidimensional Information Processing, East China Normal University \\
    $^{2}$Shanghai Changzheng Hospital, Naval Military Medical University\\
    $^{3}$Chongqing Key Laboratory of Precision Optics, Chongqing Institute of East China Normal University\\
}

\begin{document}
%
\maketitle

\begin{abstract} 
Remote photoplethysmography (rPPG) enables non-contact heart rate measurement from facial videos, but its performance is significantly degraded by facial motions such as speaking and head shaking. To address this issue, we propose two plug-and-play modules. The Angle-guided ROI Adaptive Optimization module quantifies ROI–Camera angles to refine motion-affected signals and capture global motion, while the Multi-region Joint Graph Signal Denoising module jointly models intra- and inter-regional ROI signals using graph signal processing to suppress motion artifacts. The modules are compatible with reflection model-based rPPG methods and validated on three public datasets. Results show that jointly use markedly reduces MAE, with an average decrease of 20.38\% over the baseline, while ablation studies confirm the effectiveness of each module. The work demonstrates the potential of angle-guided optimization and graph-based denoising to enhance rPPG performance in motion scenarios.

\end{abstract}
\input{Video_based_Heart_Rate_Estimation_with_Angle_guided_ROI_Optimization_and_Graph_Signal_Denoising/Sec/Introduction}

\input{Video_based_Heart_Rate_Estimation_with_Angle_guided_ROI_Optimization_and_Graph_Signal_Denoising/Sec/Methods}

\input{Video_based_Heart_Rate_Estimation_with_Angle_guided_ROI_Optimization_and_Graph_Signal_Denoising/Sec/experiment}

\input{Video_based_Heart_Rate_Estimation_with_Angle_guided_ROI_Optimization_and_Graph_Signal_Denoising/Sec/Conclusion}

\bibliographystyle{IEEEbib}
\bibliography{refs}

\end{document}

%% file: Video_based_Heart_Rate_Estimation_with_Angle_guided_ROI_Optimization_and_Graph_Signal_Denoising/Sec/Introduction.tex
\section{Introduction}
\label{sec:intro}
Remote photoplethysmography (rPPG) is a non-contact, non-invasive, and cost-efficient optical technique that derives blood volume pulse (BVP) signals from skin regions in RGB video~\cite{verkruysse2008remote}. It enables the estimation of vital signs such as heart rate (HR) and respiration, with broad applicability in telemedicine, continuous health monitoring, and driver state assessment~\cite{yang2022graph,wang2021identification}.
The fundamental principle of rPPG is to extract subtle skin chromatic variations induced by cardiac activity~\cite{mcduff2015survey}. However, these signals are exceedingly weak and susceptible to motion artifacts, which substantially constrain the robustness of rPPG-based methods in multi scenarios.

\begin{figure}[htbp]
\centering
 \includegraphics[width=0.95\linewidth]{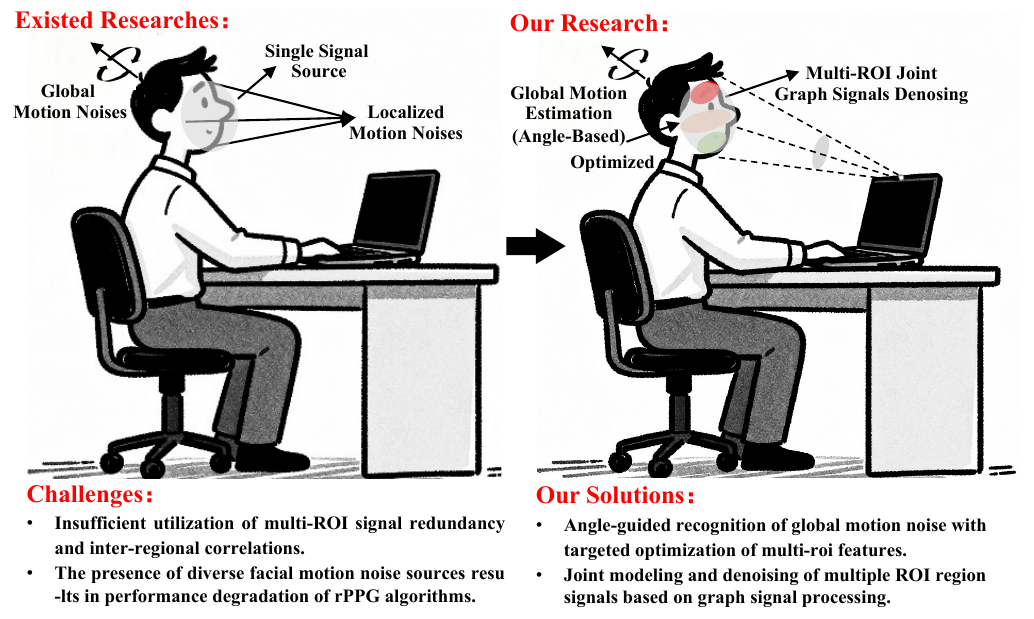} \vspace{-1.5em} 
\caption{Challenges of reflection model-based rPPG methods for HR estimation and the proposed solutions.}
\label{scene}\vspace{-1.5em} 
\end{figure}
To address the performance degradation of rPPG-based HR estimation caused by motion artifacts, existing studies have primarily followed two research directions. The first is improved optical reflection modeling~\cite{de2014improved,wang2016algorithmic,casado2023face2ppg}, which treats the color variations captured by cameras as an additive mixture of noise and BVP signals, and seeks a dedicated orthogonal projection plane to suppress noise. The second is deep learning~\cite{liu2023efficientphys,zou2025rhythmformer,yu2022physformer}, which exploits data-driven approaches to learn spatiotemporal mappings from large-scale samples, thereby enabling the automatic extraction of robust HR–related features. Despite the significant progress achieved in existing studies, most methods treat the face as a holistic signal source and extract a single aggregated signal for subsequent processing, overlooking the intrinsic physiological heterogeneity of facial regions~\cite{song2025video}. In practice, distinct facial areas exhibit different signal characteristics due to variations in localized motion disturbances, and averaging across the entire face will lead to degraded performance~\cite{cantrill2024orientation}. Identifying and integrating high-quality facial ROIs, while suppressing or attenuating the influence of low-quality regions, and further exploiting the interrelations among signals from multiple ROIs can enhance the performance~\cite{wang2014exploiting} of HR estimation algorithms.

As shown in Fig.~\ref{scene}, to characterize motion artifacts and effectively exploit the interrelations among signals from multiple ROIs, we introduce an angle-guided ROI adaptive optimization module and a multi-region joint graph signal denoising module for accurate video-based HR monitoring. The former adaptively refines multiple facial ROIs while capturing global motion characteristics of the face. The latter applies graph signal processing to jointly analyze inter- and intra-regional ROI signals, thereby suppressing motion artifacts. The contributions of this paper are as follows:
\begin{itemize}\setlength\itemsep{0pt}\setlength\parsep{0pt}\vspace{-\topsep}
    \item Proposed an angle-guided ROI adaptive optimization module, which preserves signal redundancy while adaptively refining signals affected by motion artifacts and capturing global head motion characteristics.
    \item Proposed a multi-region joint graph signal denoising module that correlates and models multiple ROI signals across different and within the same facial regions, thereby suppressing motion artifacts effectively. 
\end{itemize}

%% file: Video_based_Heart_Rate_Estimation_with_Angle_guided_ROI_Optimization_and_Graph_Signal_Denoising/Sec/Methods.tex
\section{Method}
\label{sec:method}
\begin{figure*}[htbp]
\centering
 \includegraphics[width=0.98\linewidth]{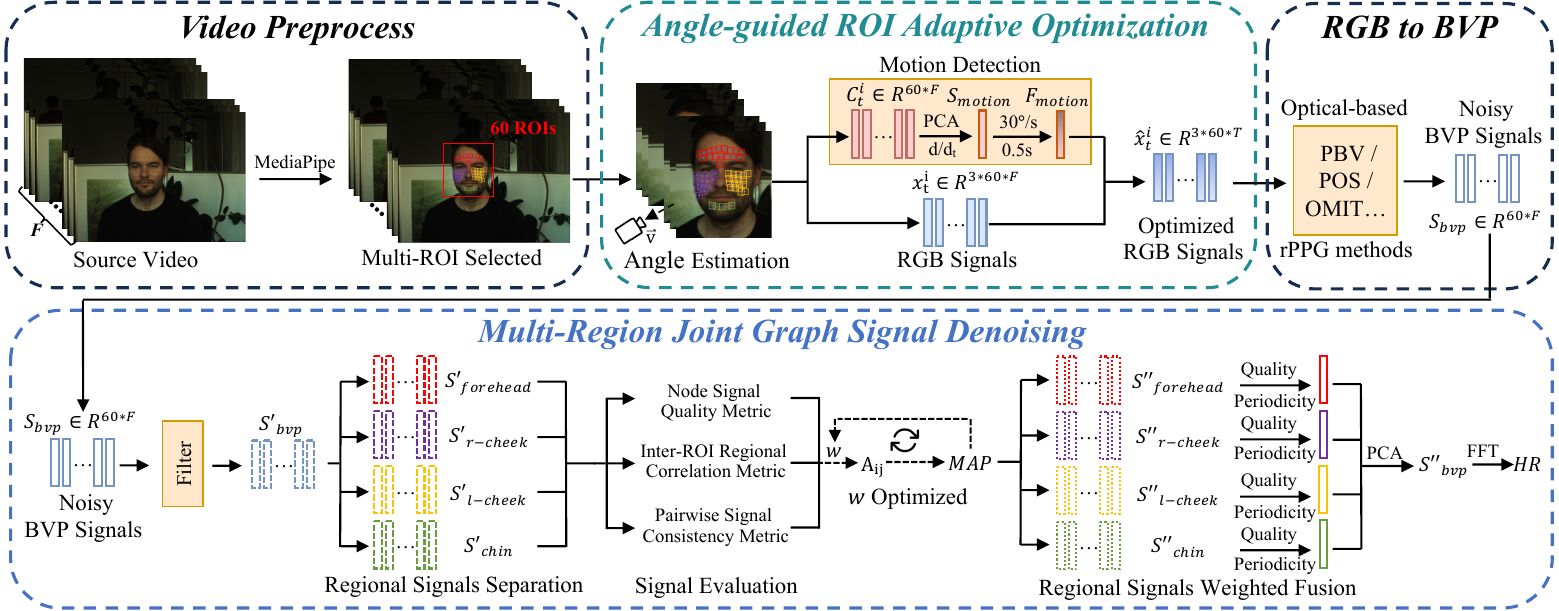}\vspace{-1.0em} 
\caption{Pipeline of HR measurement framework is shown, with variables consistent with those in Sec.~\ref{sec:method}. Angle-guided ROI Adaptive Optimization module aims to detect global head motion based on angular features and to refine ROI signals, while Multi-Region Joint Graph Signal Denoising module purifies heart rate signals by jointly modeling and denoising ROI signals.}
\label{pipeline}\vspace{-1.0em} 
\end{figure*}


\subsection{Angle-guided ROI Adaptive Optimization}
The Pipeline of the proposed method is illustrated in Fig.~\ref{pipeline}. 
MediaPipe provides dense 3D coordinates of 468 facial landmarks. Leveraging facial midline symmetry, 60 landmarks are selected on the forehead, cheeks, and chin; for each selected landmark, a $20 \times 20$-pixel rectangular ROI is constructed centered on that landmark, yielding 60 ROIs in total. 
Among the 60 ROIs, 19 are on the forehead, 18 on each cheek, and 5 on the chin. Landmark selection deliberately avoids potential occlusions. 
Since ROI signal quality degrades with orientation deviations from the camera and head motion alters ROI orientations~\cite{cantrill2024orientation,yang2017estimating}, it is necessary to quantify the angular feature $C$ for each ROI in every frame. 

The angular feature is derived from the ROI plane normal, approximated by the normal of a triangular plane defined by the ROI center and one edge.  Since only the 3D coordinates of the ROI center are available, while its four corner points $P_{n}$ are defined in 2D, their depth information must be estimated. Leveraging the dense 3D landmarks $P_{k}$ provided by MediaPipe, the depth feature of each corner point $P_{n}$ is approximated by associating it with the nearest landmark $P_{k(n)}$ in 2D space. The 2D distance are recorded as association error $e(P_n)$.
For each ROI edge, association error is defined as the sum of its two corner point association errors. The edge with the minimum error, together with the ROI center, is then used to compute the ROI plane normal. With the camera vector set to $\mathbf{v}(0,0,-1)$, the angular feature $C$ of an ROI is computed as the arccosine of the dot product between the unit normal of the triangular plane and the camera vector. 
The per-frame angular features of each ROI collectively form its angular signal. To quantify global head motion, angular signals from all 60 ROIs are fused via PCA to obtain the global angular signal, and its first-order derivative yields the motion signal $S_{\text{motion}}$.

\begin{figure}[htbp]
\centering
 \includegraphics[width=1.0\linewidth]{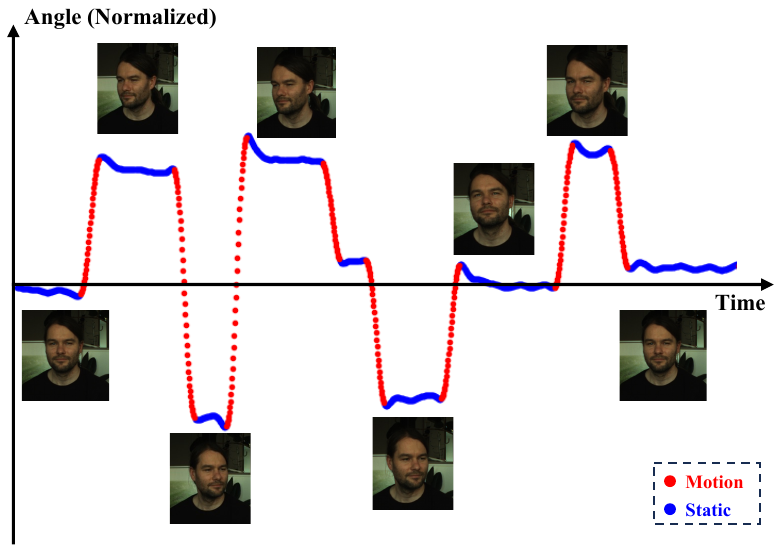} \vspace{-1.0em} 
\caption{Global motion detection can effectively identifying head rotations and nods on the PURE dataset.}
\label{motion}\vspace{-1.0em} 
\end{figure}

Motion artifacts can be categorized into global noise from rapid head rotations, which severely disrupt ROI color feature extraction, and local noise from speaking or expressions, which causes only regional disturbances. Extremely slow global motion does not affect the accuracy of rPPG-based HR estimation, as such low-frequency artifacts are eliminated by band-pass filtering.
$S_{\text{motion}}$ characterizes the rate of head angular variation and thus effectively quantifies rapid global motion in videos. After removing spurious spikes in $S_{\text{motion}}$ with a sliding mean filter, time intervals with angular velocities exceeding $30^\circ$/s for more than 0.5s are labeled as global motion sequences $F_{\text{motion}}$. Fig~\ref{motion}  shows that the method effectively identifies rapid global motion on the PURE~\cite{stricker2014non}. During rapid global head motion, substantial changes in the angle between facial ROIs and the camera lead to degraded illumination quality and reduced signal reliability.
To mitigate this issue, we replace poor-quality ROIs with their nearest high-quality neighbors. Specifically, ROIs with orientation angles less than $60^\circ$ to the camera are regarded as high quality. During video segments with global motion, the RGB features $\mathbf{x}^{i}_t$ of poor-quality ROIs are replaced by those of the nearest high-quality ROI based on center-point proximity to obtain optimized RGB feature $\tilde{\mathbf{x}}^{i}_t$.
The RGB feature sequences of the angle-guided optimized ROIs are processed with an optical reflection model-based rPPG algorithm, yielding noisy BVP signal matrix $S_{\text{bvp}} \in R^{60 \times F}$ .
\vspace{-1em}

\subsection{Multi-Region Joint Graph Signal denoising}
Prior to multi-ROI BVP graph construction, $S_{\text{bvp}}$ is band-pass filtered $[0.75-2.5]$ Hz with Butterworth filter to get $S'_{\text{bvp}}$, assigned a region label $\ell^{i}\in$\{\text{forehead}, \text{left cheek}, \text{right cheek}, \text{chin}\}, and normalized within its region to remove baseline differences to obtain 60 normalized noisy BVP signals $s^{i}_{\text{bvp}}$. 
We construct the adjacency matrix $A \in{R}^{60\times60}$ through a multidimensional weighting scheme that integrates three factors, node signal quality metric $q_{ij}$, inter-ROI regional correlation metric $r_{ij}$, and pairwise signal consistency metric $d_{ij}$.
\begin{equation}
A_{ij} = w_1 \cdot q_{ij} 
       + w_2 \cdot r_{ij} 
       + w_3 \cdot d_{ij}, 
\ i,j \in \{1,\dots,60\}, \; i \neq j
\end{equation}
$i$, $j$ denote indices of ROI nodes, $w_1$, $w_2$, $w_3$ denote the weighting coefficients of the three evaluation metrics.

$\bullet$ \textbf{Node Signal Quality metric}. 
Each node signal is evaluated by combining the power spectral ratio in the HR frequency band with inter-frame stability, and high-quality signals are selected within each region based on quality ranking. For the $i$-th node, the power spectral ratio $q^{i}_{ratio}$ within the HR frequency band is computed as follows:
\begin{equation}
q^{i}_{ratio} = 
\frac{\displaystyle \sum_{f \in [0.75,\,2.5]\,\text{Hz}} \left| \text{FFT}\!\left(s^{i}_{bvp}\right)[f] \right|^2}
     {\displaystyle \sum_{f \geq 0\,\text{Hz}} \left| \text{FFT}\!\left(s^{i}_{bvp}\right)[f] \right|^2 + 10^{-8}} .
\end{equation}
The inter-frame signal stability $q^{i}_{stability}$ is computed:
\begin{equation}
q^{i}_{stability} = 
\frac{\mu\!\left(\Delta s^{i}_{bvp}\right)}
     {\sigma\!\left(\Delta s^{\text{bvp}}_i\right)+10^{-8}}
\end{equation}
$\Delta s_{\text{bvp}}^i$ denotes the sequence of absolute differences between adjacent frames. $\sigma\left( \Delta s_{\text{bvp}}^i \right)$ represents the standard deviation of $\Delta s_{\text{bvp}}^i$, which measures the dispersion of inter-frame variations, while $\mu\left( \Delta s_{\text{bvp}}^i \right)$ represents its mean, reflecting the average amplitude of inter-frame variations. A larger $q^{i}_{\text{ratio}}$ indicates better quality, and a smaller $q^{i}_{\text{stability}}$ is preferred.
The final signal quality ranking is obtained by averaging the two rankings, and the top 50\% of signals within each region are labeled as high-quality. If two nodes belong to the same region and are both of high quality, then $q_{ij}=1.4$. If they belong to different regions, then $q_{ij}=1.1$. If either node is of low quality, then $q_{ij}=1.0$.

$\bullet$ \textbf{Inter-ROI Regional Correlation metric}.
The weighting is designed to preset the baseline connectivity strength between different facial regions based on their anatomical associations, thereby facilitating more reliable signal representation.
Moreover, spatially adjacent regions exhibit more similar blood flow variations and therefore merit higher baseline connection weights. Therefore, the inter-ROI regional correlation weight $r_{ij}$ is following the rules:
\begin{equation}
r_{ij} =
\begin{cases}
1.0, & \ell^i = \ell^j, \\
0.6, & \{\ell^i, \ell^j\} = \{\text{cheek}, \text{forehead}\}, \\
0.4, & \{\ell^i, \ell^j\} = \{\text{cheek}, \text{chin}\}, \\
0.2, & \{\ell^i, \ell^j\} = \{\text{forehead}, \text{chin}\}.
\end{cases}
\end{equation}
Note that the left and right cheeks are treated as the same region, 
thus their pairwise similarity is also set to $1.0$.

$\bullet$ \textbf{Pairwise Signal Consistency metric}.
The weight is defined as the dynamic synchronization between two node signals, quantified via sliding-window correlation. A window of 0.8\,s (approximately one cardiac cycle) with a 0.4\,s step is used, and the mean Pearson correlation across windows is taken as the consistency parameter. To account for signal quality, consistencies involving low-quality nodes are down-weighted. Specifically, the final inter-node consistency equals the consistency parameter multiplied by a quality weight of $1.0$ for high-quality nodes and $0.6$ for low-quality nodes.
We employs a maximum a posteriori (MAP) estimation method for graph signal denoising. First, based on the obtained adjacency matrix $A$ to constructe the degree matrix $D_{ii} = \sum_{j=1}^{60}A_{ij}$. Therefore, a combinatorial graph Laplacian can be expressed as $L=D-A$. Therefore, the MAP problem to be solved can be formulated as:
\begin{equation}
\arg\min_{\mathbf{S''_{bvp}}} \; \| \mathbf{S'_{bvp}} - \mathbf{S''_{bvp}} \|_2^2 \;+\; \lambda \, \mathbf{S''_{bvp}}^\top L \mathbf{S''_{bvp}}\label{map}
\end{equation}
\(\mathbf{S'_{bvp}}\) denotes the observed noisy graph signal, \(\mathbf{S''_{bvp}}\) the desired clean signal, and \(\lambda\) is regularization parameter. Since the objective function needs to be minimized, the analytical solution can be obtained by setting its derivative to zero. 

The MAP estimator admits a closed-form solution $\mathbf{\hat{S}''_{bvp}} = (I + \lambda L)^{-1}\mathbf{S'_{bvp}}$, where \(I\) is identity matrix. Since \(\lambda>0\), it follows that $I + \lambda L >0$, and thus $I + \lambda L$ is invertible, ensuring that the solution of Eq.~\ref{map} is stable. Since $L$ depends on $A_{ij}$, and the weighting coefficients $w=\{w_1, w_2, w_3\}$ are unknown, it is necessary to adaptively determine suitable coefficient values. The weight estimation is essentially a constrained maximization problem, formulated as:
\begin{equation}
\mathbf{w}^* = \arg\max_{\mathbf{w}} \; \text{MAP}(\mathbf{w})
\quad w_1 + w_2 + w_3 = 1,\; w_k>0
\end{equation}
Since the objective function $\text{MAP}(\mathbf{w})$ is nonlinear, non-convex, and lacks an explicit analytical form, it is solved using the gradient descent method. First, the unconstrained variables $\mathbf{z} = \{z_1, z_2, z_3\}$ are transformed into constraint-satisfying weights through a softmax mapping:
\begin{equation}
w_k = \frac{\exp(z_k)}{\sum_{m=1}^3 \exp(z_m)}, \quad k = 1,2,3
\end{equation}
At this stage, the optimization variables are replaced by $\mathbf{z}$, and the objective function is reformulated as $\text{MAP}(\mathbf{z})$. The initialization is set as $\mathbf{z}^{(0)} = \{0,0,0\}$, corresponding to the initial weights $\mathbf{w}^{(0)} = \{1/3, 1/3, 1/3\}$. The learning rate is chosen as $\eta$ = 0.01. Since $\text{MAP}(\mathbf{z})$ has no explicit gradient, numerical approximation is adopted during the iterations. Specifically, for each $z_k$, a small perturbation $\epsilon = 10^{-5}$ is added, and the gradient is approximated as:
\begin{equation}
\frac{\partial \text{MAP}}{\partial z_k} \approx
\frac{\text{MAP}(\mathbf{z} + \epsilon \cdot \mathbf{e}_k) - \text{MAP}(\mathbf{z})}{\epsilon}
\end{equation}
where $\mathbf{e}_k$ denotes the unit vector with 1 at the $k$-th position. The variable $\mathbf{z}$ is updated along the gradient direction as:
\begin{equation}
\mathbf{z}^{(t+1)} = \mathbf{z}^{(t)} + \eta \cdot \nabla_{\mathbf{z}} \text{MAP}\big(\mathbf{z}^{(t)}\big)
\end{equation}
The iteration is terminated once the gradient converges or the maximum of 30 iterations is reached, after which the final weight vector $\mathbf{w}^*$ is obtained. Graph signal denoising exploits the redundancy among multiple ROI to suppress noise, yielding 60 outputs.
The signals are first aggregated into regional signals via weighted fusion, where the weights are determined by the normalized product of signal periodicity and quality labels. Features corresponding to global motion periods are then removed, and the remaining segments are concatenated and fused using PCA to derive the final BVP signal.

\input{Video_based_Heart_Rate_Estimation_with_Angle_guided_ROI_Optimization_and_Graph_Signal_Denoising/Tab/Tab}

%% file: Video_based_Heart_Rate_Estimation_with_Angle_guided_ROI_Optimization_and_Graph_Signal_Denoising/Tab/Tab.tex
\begin{table}[!t]
\normalsize
\caption{Performance comparison of baseline rPPG methods with angle-guided ROI module (AGROI) and multi-region joint graph signal denoising module (GSD) on three public datasets. S-E: Signal Extraction; S-P: Signal Processing; SROI: Single ROI; BP: Bandpass; BGSD: Bandpass \& GSD.}
\setlength{\tabcolsep}{1mm}
    \centering
    \renewcommand{\arraystretch}{1.0} 
   \resizebox{0.5\textwidth}{!}{
    \begin{tabular}{c|c|cc|cc|cc}
    \toprule
   \multirow{2.5}{*}{\textbf{S-E}} &\multirow{2.5}{*}{\textbf{S-P}} & \multicolumn{2}{c|}{\textbf{COHFACE} }& \multicolumn{2}{c|}{\textbf{PURE} } & \multicolumn{2}{c}{\textbf{MMPD} } \\
    \cmidrule{3-8}
   &  & \textbf{MAE↓} & \textbf{MAPE↓} & \textbf{MAE↓} & \textbf{MAPE↓} & \textbf{MAE↓} & \textbf{MAPE↓} \\
    \midrule
     PBV \& SROI &  BP&7.61&11.83&6.54&8.15&20.40&22.35\\
    PBV \& AGROI&BP &5.82&\textbf{6.17}&5.72&6.24&19.87&21.73\\
  \textbf{PBV \& AGROI} &\textbf{BGSD}&\textbf{5.68}&8.29&\textbf{4.52}&\textbf{6.13}&\textbf{18.94}&\textbf{21.59}\\
   \midrule
    POS \& SROI &  BP&5.18&7.18&5.88&6.33&16.16&18.06\\
    POS \& AGROI&BP &4.63&6.42&5.31&\textbf{5.46}&15.45&17.56\\
  \textbf{POS \& AGROI} &\textbf{BGSD}&\textbf{4.06}&\textbf{5.77}&\textbf{4.21}&6.53&\textbf{13.79}&\textbf{15.62}\\
     \midrule
     OMIT \& SROI &  BP&4.76&6.51&5.32&5.27&17.97&19.32\\
    OMIT \& AGROI&BP &4.33&6.27&4.99&\textbf{4.82}&17.66&18.79\\
  \textbf{OMIT \& AGROI} &\textbf{BGSD}&   \textbf{3.84}&\textbf{5.41}&\textbf{3.78}&5.47&\textbf{16.72}&\textbf{18.54}\\
     \bottomrule
    \end{tabular}
    \vspace{-1em}
    }\label{tab}
\end{table}

%% file: Video_based_Heart_Rate_Estimation_with_Angle_guided_ROI_Optimization_and_Graph_Signal_Denoising/Sec/experiment.tex
\section{Experiment and Results}
\label{sec:Experiment}

Since the modules essentially target the pre- and post-processing stages of HR estimation, their effectiveness is validated by integrating them with three mainstream reflection model-based rPPG methods (PBV~\cite{de2014improved}, POS~\cite{wang2016algorithmic}, and OMIT~\cite{casado2023face2ppg}). Experiments are conducted on COHFACE~\cite{heusch2017reproducible}, PURE~\cite{stricker2014non}, and MMPD~\cite{tang2023mmpd} using MAE and MAPE as evaluation metrics. The baseline is self-developed, while the PBV, POS, and OMIT codes are adapted from rPPG-Toolbox~\cite{liu2023rppg}. 
Tab.~\ref{tab} presents the experimental results. For each reflection model-based rPPG method, the baseline uses the entire face as a single ROI, with the extracted BVP signal post-processed only by Butterworth band-pass filter in the HR frequency band. In the second row of each group, signal extraction is performed with the Angle-guided ROI Adaptive Optimization module, while processing still relies solely on the Butterworth filter: multiple optimized ROI signals are first obtained, individually filtered, and then fused using PCA. The third row corresponds to the full configuration, where both proposed modules are applied. Results show that combining the two modules significantly improves MAE, indicating enhanced HR estimation accuracy. Moreover, comparison of the second and third rows validates the individual effectiveness of each module. 

%% file: Video_based_Heart_Rate_Estimation_with_Angle_guided_ROI_Optimization_and_Graph_Signal_Denoising/Sec/Conclusion.tex
\section{Conclusion}
\label{sec:conclusion}
In this work, we propose two plug-and-play modules that an Angle-guided ROI Adaptive Optimization module and a Multi-region Joint Graph Signal Denoising module to enhance video-based HR estimation. The first module adaptively refines ROI signals by leveraging head pose variations while preserving signal redundancy, and the second exploits inter- and intra-regional correlations to effectively suppress motion-induced noise through graph signal processing. Experimental results on three public datasets demonstrate that the joint application of the two modules consistently improves performance, thereby enhances HR estimation accuracy, offering a robust and generalizable solution for non-contact physiological monitoring in real-world scenarios.
